\pdfoutput=1
\documentclass[11pt,a4paper]{article}
\usepackage{times}
\usepackage{latexsym}
\usepackage{url}

\usepackage{mathtools}
\usepackage{setspace} 
\usepackage{dsfont}
\usepackage{amsfonts}

\usepackage{amssymb}

\usepackage{amsmath}
\DeclareMathOperator*{\argmax}{argmax}

\DeclareMathOperator{\EX}{\mathbb{E}}

\usepackage{subcaption}
\usepackage{paralist}
\usepackage{times}
\usepackage{latexsym}
\usepackage{graphicx}
\usepackage{booktabs}
\usepackage{numprint}
\usepackage[T1]{fontenc}
\usepackage{tikz}
\usepackage{url}
\usepackage{pgfplotstable}
\usepackage{titlesec}
\usepackage{color}
\usepackage{lipsum,adjustbox}
\usepackage[font={small}]{caption}
\usetikzlibrary{positioning}
\usepackage{bbm}
\usepackage{multirow}

\makeatletter
\newcommand{\@BIBLABEL}{\@emptybiblabel}
\newcommand{\@emptybiblabel}[1]{}

\usepackage[hyperref]{emnlp-ijcnlp-2019}
\graphicspath{{./plots/}}
\newcommand{\com}[1]{}

\newcommand{\pke}[0]{{\sc PkE}}

\newenvironment{myequation}{
	\vspace{-1em}
	\begin{equation}
}{
	\end{equation}
	\vspace{-1.2em}
}
\newenvironment{myequation*}{
	\vspace{-1em}
	\begin{equation*}
}{
	\end{equation*}
	\vspace{-1.2em}
}

\aclfinalcopy
\usepackage{xr}
\begin{document}
	\title{On the Weaknesses of Reinforcement Learning \\
		for Neural Machine Translation}

	\author{
		Leshem Choshen\textsuperscript{1} \And Lior Fox\textsuperscript{1}	
		 \And Zohar Aizenbud\textsuperscript{1}\\
		 \hspace{-4.5cm}\textsuperscript{1}School of Computer Science and Engineering,
		 \textsuperscript{2} Department of Cognitive Sciences \\
		 \hspace{-4.5cm}The Hebrew University of Jerusalem \\
		 \hspace{-4.5cm}\texttt{\{leshem.choshen, lior.fox, zohar.aizenbud\}@mail.huji.ac.il}\\ \hspace{-4.5cm}\texttt{oabend@cs.huji.ac.il}\\ 
		 \And Omri Abend\textsuperscript{1,2} \\		
	}
	\maketitle
	
	\begin{abstract}
Reinforcement learning (RL) is frequently used to increase performance in text generation tasks,
including machine translation (MT), 
notably through the use of Minimum Risk Training (MRT) and Generative Adversarial Networks (GAN). 
However, little is known about what and how these methods learn in the context of MT. 
We prove that one of the most common RL methods for MT does not optimize the 
expected reward, as well as show that other methods take an infeasibly long time to converge.
In fact, our results suggest that RL practices in MT are likely to improve performance
only where the pre-trained parameters are already close to yielding the correct translation.
Our findings further suggest that observed gains may be due to effects unrelated to the training signal,
but rather from changes in the shape of the distribution curve.
	\end{abstract}
	
	\section{Introduction}
	
	Reinforcement learning (RL) is an appealing path for advancement in Machine Translation (MT),
	as it allows training systems to optimize non-differentiable score functions, 
	common in MT evaluation, as well as its ability to tackle the ``exposure bias'' \citep{ranzato2015sequence}
        in standard training, namely that the model is not exposed during training to incorrectly generated tokens, 
	and is thus unlikely to recover from generating such tokens at test time. 
	These motivations have led to much interest in RL for text generation in general and MT in particular.
	Various policy gradient methods have been used, notably {\sc Reinforce} \citep{williams1992simple} and variants thereof 
	\citep[e.g.,][]{ranzato2015sequence,edunov2018classical} and Minimum Risk Training \citep[MRT; e.g.,][]{och2003minimum, shen2016minimum}.
	Another popular use of RL is  for training GANs  \cite{yang2018improving,tevet2018evaluating}. 
	See \S\ref{sec:rl_for_generation}.
	Nevertheless, despite increasing interest and strong results, 
	little is known about what accounts for these performance gains, 
	and the training dynamics involved.

        We present the following contributions.
        First, our theoretical analysis shows that commonly used approximation methods are theoretically ill-founded,
	and may converge to parameter values that do not minimize the risk, nor are local minima thereof (\S\ref{sec:mrt_theory}).

        Second,
        using both naturalistic experiments and carefully constructed simulations,
        we show that performance gains observed in the literature likely stem not 
        from making target tokens the most probable, but from unrelated effects, such as
        increasing the \emph{peakiness} of the output distribution (i.e., the probability mass of
        the most probable tokens). We do so by comparing a setting where the reward is informative,
        vs. one where it is constant. In \S\ref{sec:pke} we discuss this peakiness effect (\pke).

        Third, we show that promoting the target token to be the mode is likely to take a prohibitively long time.
	The only case we find, where improvements are likely, is where the target token is among the first 2-3
        most probable tokens according to the pretrained model.
	These findings suggest that {\sc Reinforce} (\S\ref{sec:convergence_rate}) and CMRT (\S\ref{sec:MRT_exp})
	are likely to improve over the pre-trained model only under the best possible conditions, i.e.,
        where the pre-trained model is ``nearly'' correct.  
          

	We conclude by discussing other RL practices in MT which should be avoided for
    practical and theoretical reasons, and briefly discuss alternative RL approaches that
    will allow RL to tackle a larger class of errors in pre-trained models (\S\ref{sec:discussion}).
	

	\section{RL in Machine Translation}\label{sec:rl_for_generation}
	
	\subsection{Setting and Notation} 
	
	An MT system generates tokens $y=\left(y_1,...,y_n\right)$ 
	from a vocabulary $V$ one token at a time.
	The probability of generating $y_i$ given preceding tokens $y_{<i}$ is given by $P_{\theta}(\cdot|x,y_{<i})$, where
	$x$ is the source sentence and $\theta$ are the model parameters. 
	For each generated token $y$, we denote with $r(y; y_{<i},x,y^{(ref)})$ the score, or reward, for generating $y$ 
	given $y_{<i}$, $x$, and the reference sentence $y^{(ref)}$.
	For brevity, we omit parameters where they are fixed within context.
	For simplicity, we assume $r$ does not depend on following tokens $y_{>i}$.

        We also assume there is exactly one valid target token, as in practice MT systems are trained against a single
        reference translation per sentence \cite{Schulz2018ASD}. 
	In practice, either a token-level reward is approximated using Monte-Carlo methods  \citep[e.g.,][]{yang2018improving}, 
	or a sentence-level (sparse) reward is given at the end of the episode (sentence). 
	The latter is equivalent to a uniform token-level reward. 

	
	$r$ is often either the negative log-likelihood, or based on
	standard MT metrics, e.g., BLEU \citep{papineni2002bleu}.
	When applying RL in MT, we seek to maximize the expected reward (denoted with $R$); i.e., to find
	
	\begin{myequation}
		\theta^* = \argmax_\theta\,\, R(\theta) = \argmax_\theta\,\, \EX_{y \sim P_\theta}[r(y)]
		\label{eq:r}
	\end{myequation} 
	\vspace{-0.3cm}
	
	\subsection{\sc Reinforce}
	
	For a given source sentence, and partially generated sentence $y_{<i}$,
	{\sc Reinforce} \citep{williams1992simple} samples $k$ tokens ($k$ is a hyperparameter) $S=\left(y^{(1)},...,y^{(k)}\right)$ from $P_\theta$ and updates $\theta$ according
	to this rule:
	
	\begin{myequation}
		\Delta\theta \propto \frac{1}{k}\sum_{i=1}^k r(y_i)\nabla\log(P_\theta(y_i)) 
		\label{eq:reinforce}
	\end{myequation}
	\vspace{.1cm}
	
	The right-hand side of Eq. \eqref{eq:reinforce} is an unbiased estimator of the gradient of the objective function, i.e., $\mathbb{E}\left[\Delta \theta\right] \propto \nabla_\theta R\left(\theta\right)$. Therefore, {\sc Reinforce} is performing a form of stochastic gradient ascent on $R$, and has similar formal guarantees. From here follows that if $R$ is constant with respect to $\theta$, then the expected $\Delta\theta$ prescribed by {\sc Reinforce} is zero. We note that $r$ may be shifted by a constant term (called a ``baseline''), without affecting the optimal value for $\theta$. 
	
	
	{\sc Reinforce} is used by a variety of works in MT, text generation, and image-to-text tasks \citep{liu2016optimization,Wu2018ASO,rennie2017self, shetty2017speaking,Hendricks2016Generating} -- 
	in isolation, or as a part of training \cite{ranzato2015sequence}.
	Lately, an especially prominent use for {\sc Reinforce} is adversarial training with discrete data, where another network predicts the reward (GAN). 
	For some recent work on RL for NMT, see \citep{zhang2016generating,li2017adversarial,wu2017adversarial,yu2017seqgan,yang2018improving}.
	
	\subsection{Minimum Risk Training}\label{sec:mrt_theory}
	
	The term Minimum Risk Training (MRT) is used ambiguously in MT to refer either to the application of {\sc Reinforce} to
	minimizing the risk (equivalently, to maximizing the expected reward, the negative loss),
	or more commonly to a somewhat different estimation method, which we term Contrastive MRT (CMRT) and turn now to analyzing. 
	CMRT was proposed by \citet{och2003minimum}, adapted to NMT by \citet{shen2016minimum},
	and often used since \citep{ayana2016neural, neubig2016Lexicons, shen2017Optimizing, edunov2018classical, makarov2018Neural, neubig2018XNMTTE}.
	
	
	The method works as follows: at each iteration, sample $k$ tokens $S = \{y_1,\ldots, y_k\}$ from $P_{\theta}$, 
	and update $\theta$ according to the gradient of
	\vspace{-.3cm}
	
	\begin{myequation*}
		\widetilde{R}(\theta,S) = \sum_{i=1}^k Q_{\theta,S}(y_i)r(y_i) = \EX_{y\sim Q}\big[r(y)\big]
	\end{myequation*}
	
	\noindent  where
	\vspace{-.5cm}
	
	\begin{myequation*}
		Q_{\theta,S}(y_i) = \frac{P(y_i)^{\alpha}}{\sum_{y_j \in S} P(y_j)^{\alpha}}
	\end{myequation*}
	\vspace{.1cm}
	
	Commonly (but not universally), deduplication is performed, so $\widetilde{R}$ sums over a set of unique values \cite{sennrich2017nematus}.  
	This changes little in our empirical results and theoretical analysis.
	
	Despite the  resemblance in definitions of $R$ (Eq. \eqref{eq:r}) and $\widetilde{R}$ (indeed, $\widetilde{R}$ is sometimes presented as an approximation of $R$), 
	they differ in two important aspects.  
	First, $Q$'s support is $S$, so increasing $Q(y_i)$ for some $y_i$ necessarily comes at the expense
	of $Q(y)$ for some $y\in S$. In contrast, increasing $P(y_i)$, as in {\sc Reinforce}, may come at the expense of 
	$P(y)$ for any $y \in V$.
	Second, $\alpha$ is a smoothness parameter: the closer $\alpha$ is to 0, the closer $Q$ is to be uniform.
	
	We show in Appendix \ref{app:counter_ex} that despite its name, CMRT does not optimize $R$, nor does it optimize  $\EX[\widetilde{R}]$.
        That is, it may well converge to values that are not local maxima of $R$, making it theoretically ill-founded.\footnote{\citet{sakaguchi2017grammatical} discuss the relation between CMRT and {\sc Reinforce}, claiming that CMRT is a variant of {\sc Reinforce}. Appendix \ref{app:counter_ex} shows that CMRT does not in fact optimize the same objective.}
	However, given that CMRT is often used in practice, the strong results it yielded and the absence of theory for explaining it,
	we discuss it here.
	Given a sample $S$, the gradient of $\widetilde{R}$ is given by
	\begin{myequation}\label{eq:Rtilde_grad}
		\begin{split}
			\nabla \widetilde{R} =& \alpha \sum_{i=1}^k \Big( Q(y_i) \cdot r(y_i) \cdot \nabla\log P(y_i) \Big) \\&- \EX_{Q}[r]\nabla\log Z(S)
		\end{split}
	\end{myequation}
	
	\vspace{.2cm}
	\noindent
	where $Z(S) = \sum_i P(y_i)^{\alpha}$. See Appendix \ref{app:grad_mrt}.
	
	Comparing Equations \eqref{eq:reinforce} and \eqref{eq:Rtilde_grad}, the differences between {\sc Reinforce} and CMRT are reflected again. 
	First, $\nabla\widetilde{R}$ has an additional term, proportional to $\nabla\log{}Z(S)$, which yields the contrastive effect.
	This contrast may improve the rate of convergence since it counters the decrease of probability mass for non-sampled tokens.
	
	Second, for a given $S$, the relative weighting of the gradients $\nabla\log{}P(y_i)$ is proportional to $r(y_i)Q(y_i)$, or equivalently to $r(y_i)P(y_i)^\alpha$.
	CMRT with deduplication sums over distinct values in $S$ (Eq. \eqref{eq:Rtilde_grad}), while {\sc Reinforce} sums over all values. This means that the relative weight of the unique value $y_i$ is $\frac{r(y_i)|\{y_i \in S\}|}{k}$ in {\sc Reinforce}. For $\alpha=1$ the expected value of these relative weights is the same, and so for $\alpha < 1$ (as is commonly used), more weight is given to improbable tokens, which could also have a positive effect on the convergence rate.\footnote{Not performing deduplication results in assigning higher relative weight to high-probability tokens, which may have an adverse effect on convergence rate. For an implementation without deduplication, see THUMT \cite{zhang2017thumt}.}
	However, if $\alpha$ is too close to 0, $\nabla\widetilde{R}$ vanishes.
	This tradeoff explains the importance of tuning $\alpha$ reported in the literature.
	In \S\ref{sec:MRT_exp} we present simulations with CMRT, showing very similar trends as presented by {\sc Reinforce}.
	
	%
	%
	%
	%
	\section{Motivating Discussion}\label{sec:motivation}
	
	Implementing a stochastic gradient ascent, {\sc Reinforce} is guaranteed to converge to a stationary point of $R$ under broad conditions.
	However, not much is known about its convergence rate under the prevailing conditions in NMT.
	
	We begin with a qualitative, motivating analysis of these questions.
	As work on language generation empirically showed, RNNs quickly learn to output very peaky distributions \cite{press2017language}. 
	This tendency is advantageous for generating fluent sentences with high probability, but may also entail slower convergence rates when using RL to fine-tune the model,
	because RL methods used in text generation sample from the (pretrained) policy distribution, which means they mostly sample what the pretrained model deems to be likely. 
	Since the pretrained model (or policy) is peaky, exploration of other potentially more rewarding tokens will be limited, hampering convergence. 
	
	
	Intuitively, {\sc Reinforce} increases the probabilities of successful (positively rewarding) observations, weighing updates by how rewarding they were. 
	When sampling a handful of tokens in each context (source sentence $x$ and generated prefix $y_{<i}$), and where the number of epochs is not large, 
	it is unlikely that more than a few unique tokens will be sampled from  $P_{\theta}(\cdot|x,y_{<i})$.
	(In practice, $k$ is typically between 1 and 20, and the number of epochs between 1 and 100.)
	It is thus unlikely that anything but the initially most probable candidates will be observed. 
	Consequently, {\sc Reinforce} initially raises their probabilities, 
	even if more rewarding tokens can be found down the list. 
	
	We thus hypothesize the peakiness of the distribution, i.e., the probability mass allocated to the most probable tokens,
	will increase, at least in the first phase. 
	We call this the peakiness-effect (\pke), and show
	it occurs both in simulations (\S\ref{sec:peakiness_simulations}) and in full-scale NMT experiments (\S\ref{sec:peakiness_nmt}).
	
	With more iterations, the most-rewarding tokens will be eventually sampled, 
	and gradually gain probability mass. 
	This discussion suggests that training will be extremely sample-inefficient. We assess the rate of convergence empirically in \S\ref{sec:convergence_rate}, finding this to be indeed the case.
	
	\vspace{-0.1cm}
	\section{The Peakiness Effect}\label{sec:pke}
	\vspace{-0.1cm}
	
	We turn to demonstrate that the initially most probable tokens will initially gain probability mass, even if they are not the most rewarding, yielding a \pke.

        \citet{caccia2018language} recently observed in the context of language modeling using GANs that
        performance gains similar to those GAN yield can be achieved by decreasing the temperature
        for the prediction softmax (i.e., making it peakier).
	However, they proposed no account as to what causes this effect. Our findings propose an underlying mechanism leading to this
        trend. We return to this point in \S\ref{sec:discussion}. Furthermore, given their findings, it is reasonable to assume that our
        results are relevant for RL use in other generation tasks, whose output space too is discrete, high-dimensional and concentrated. 

	\vspace{-0.1cm}
	\subsection{Controlled Simulations}\label{sec:peakiness_simulations}
	
	We experiment with a 1-layer softmax model, that predicts a single token $i \in V$ with probability $\frac{e^{\theta_i}}{\sum_je^{\theta_j}}$.
	$\theta=\{\theta_j\}_{j\in V}$ are the model's parameters.
	
	This model simulates the top of any MT decoder that ends with a softmax layer, as essentially all NMT decoders do. 
	To make experiments realistic, we use similar parameters as those reported in the influential Transformer NMT system \citep{vaswani2017attention}.
	Specifically, the size of $V$ (distinct BPE tokens) is 30715, and the initial values for $\theta$ were sampled from 1000 sets of logits taken from  
	decoding the standard newstest2013 development set, using a pretrained Transformer model. The model was pretrained on WMT2015 training data \cite{bojar2015Findings}. 
	Hyperparameters are reported in Appendix \ref{ap:technichalities}. We define one of the tokens in $V$ to be the target token and denote it with $y_{best}$. We assign deterministic token reward, this makes learning easier than when relying on approximations and our predictions optimistic.
	We experiment with two reward functions: 
	
	\vspace{-.2cm}
	\begin{enumerate}
		\item
		{\bf Simulated Reward:} $r(y)=2$ for $y=y_{best}$, $r(y)=1$ if $y$ is one of the 10 initially highest scoring tokens, and $r(y)=0$ otherwise. 
		This simulates a condition where the pretrained model is of decent but sub-optimal quality. 
		$r$ here is at the scale of popular rewards used in MT, such as GAN-based rewards or BLEU (which are between 0 and 1).
		\item   \vspace{-.3cm}
		{\bf Constant Reward:} $r$ is constantly equal to 1, for all tokens. This setting is aimed to confirm that \pke\ is not a result of the signal carried by the reward.  
	\end{enumerate}
	\vspace{-.2cm}
	
	Experiments with the first setting were run 100 times, each time for 50K steps, updating $\theta$ after each step.
	With the second setting, it is sufficient to take a single step at a time, as the expected update after each step is zero,
	and so any \pke\ seen in a single step is only accentuated in the next. It is, therefore, more telling to run more repetitions rather than more steps per initialization. 
	We, therefore, sample 10000 pretrained distributions, and perform a single {\sc Reinforce} step.    
	
	As RL training in NMT lasts about 30 epochs before stopping, samples about 100K tokens per epoch, and as the network already predicts $y_{best}$ in about two thirds of the contexts,\footnote{Based on our NMT experiments, which we assume to be representative of the error rate of other NMT systems.} we estimate the number of steps used in practice to be in the order of magnitude of 1M.
	For visual clarity, we present figures for 50K-100K steps. 
	However, full experiments (with 1M steps) exhibit similar trends: where {\sc Reinforce} was not close to converging after 50K steps, the same was true after 1M steps.
	
	We evaluate the peakiness of a distribution in terms of the probability of the most probable token (the mode), the total probability of the ten most probable tokens, and the entropy of the distribution (lower entropy indicates more peakiness).
	
	
	\begin{figure*}[t]
		\begin{subfigure}[b]{0.5\textwidth}
			\includegraphics[width=\columnwidth]{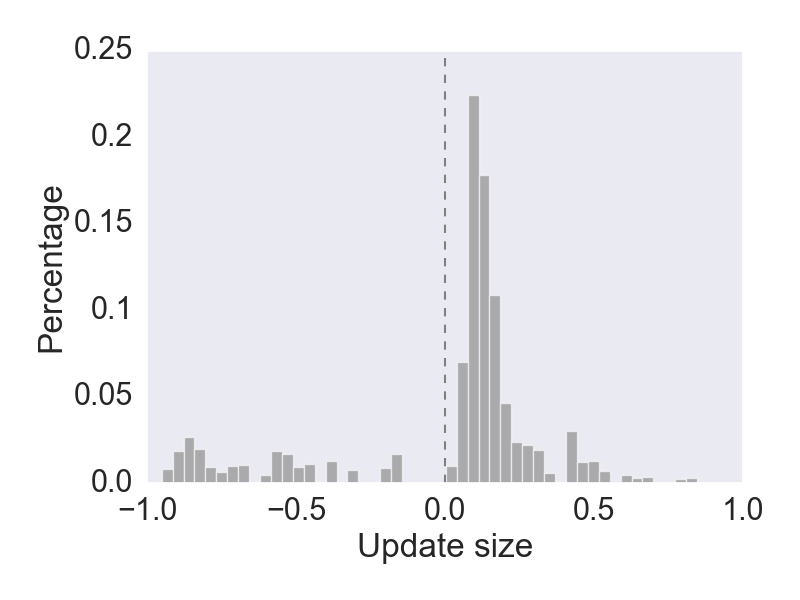}
			\caption{Top 10 \label{fig:peakiness_top10_no_reward}}
		\end{subfigure} 
		\begin{subfigure}[b]{0.5\textwidth}
			\includegraphics[width=\columnwidth]{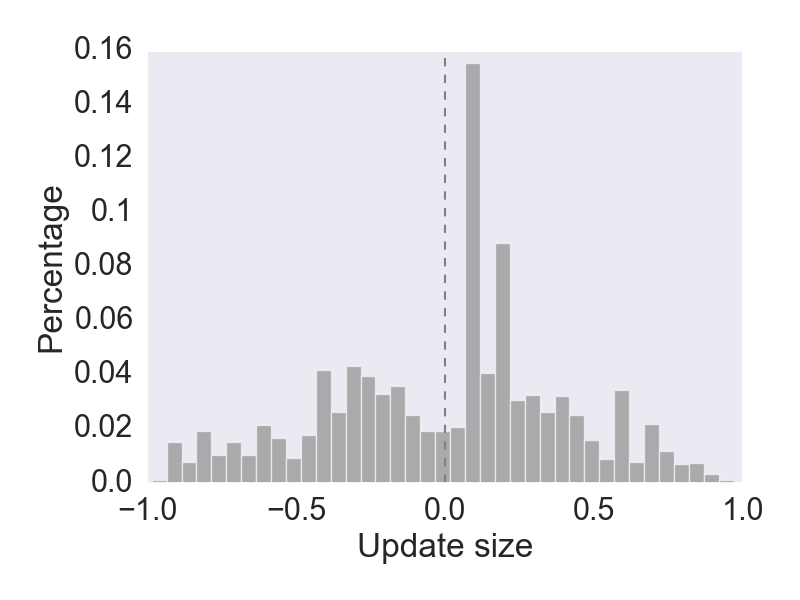}
			\caption{Mode \label{fig:peakiness_mode_no_reward}}
		\end{subfigure}
		\caption{A histogram of the update size (x-axis) to the total probability of the 10 most probable tokens (left) or the most probable token (right) 
			in the Constant Reward setting.
			An update is overwhelmingly more probable to increase this probability than to decrease it. \label{fig:peakiness}}
		\vspace{-0.3cm}
	\end{figure*}
	
	\begin{figure*}
		\begin{subfigure}[b]{0.33\textwidth}
			\includegraphics[width=\columnwidth]{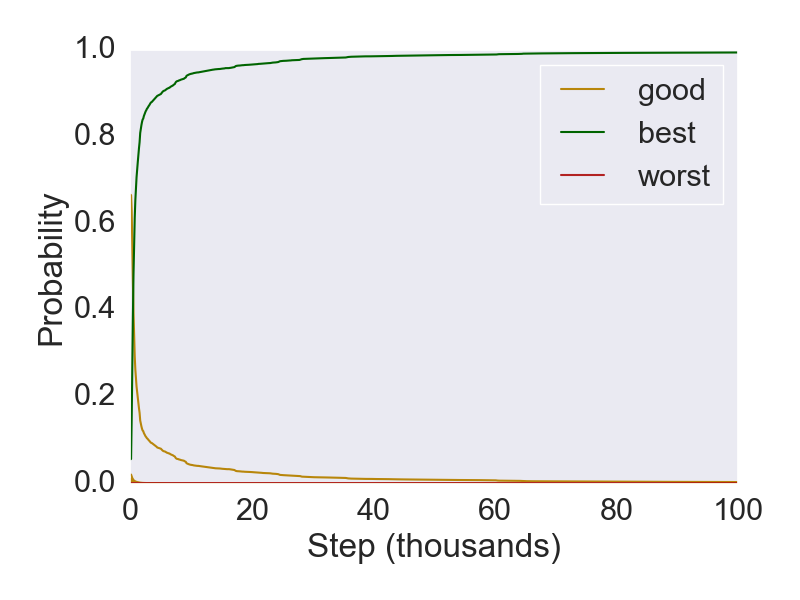}
			\caption{\label{fig:reinforce_best1}}
		\end{subfigure}
		\begin{subfigure}[b]{0.33\textwidth} 
			\includegraphics[width=\columnwidth]{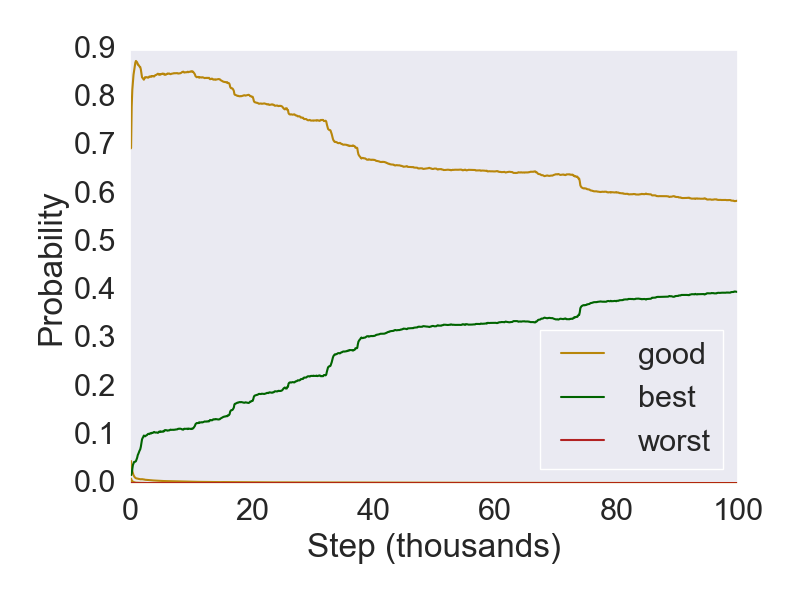}
			\caption{ \label{fig:reinforce_best2}}
		\end{subfigure}
		\begin{subfigure}[b]{0.33\textwidth} 
			\includegraphics[width=\columnwidth]{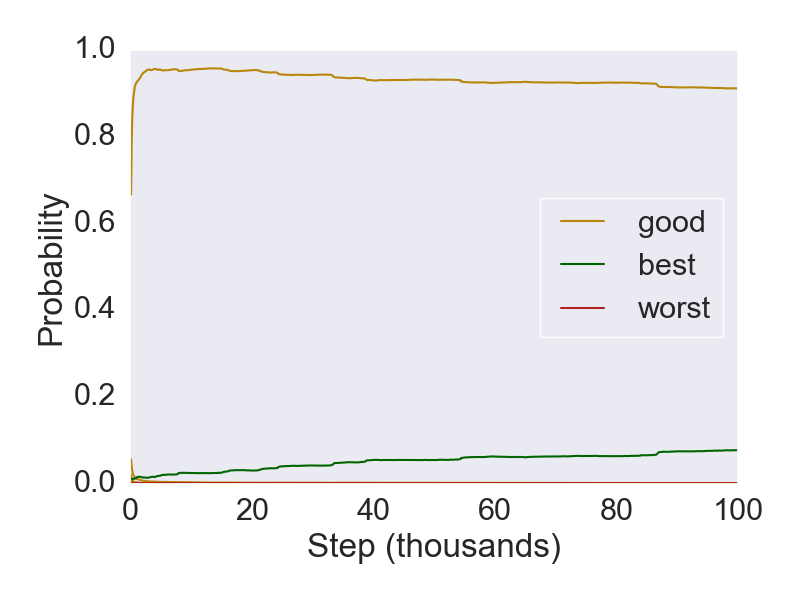}
			\caption{ \label{fig:reinforce_best3}}
		\end{subfigure}
		\caption{Token probabilities through {\sc Reinforce} training, in the controlled simulations in the Simulated Reward setting.
			The left/center/right figures correspond to simulations where the target token ($y_{best}$) was initially the second/third/fourth
			most probable token. The green line corresponds to the target token, yellow lines to medium-reward tokens and red lines to no-reward tokens.\label{fig:reinforce_3best}}
		\vspace{-0.4cm}
	\end{figure*} 
	
	\paragraph{Results.} 
	The distributions become peakier in terms of all three measures: on average, the mode's probability and the 10 most probable tokens increases,
	and the entropy decreases. Figure \ref{fig:peakiness_top10_no_reward} presents the histogram of the update size, the difference in the probability of the 10 most probable tokens in the Constant Reward setting, after a single step. Figure \ref{fig:peakiness_mode_no_reward} depicts similar statistics for the mode.
	The average entropy in the pretrained model is 2.9 is reduced to 2.85 after one {\sc Reinforce} step.
	
	Simulated Reward setting shows similar trends. For example, entropy decreases from 3 to about 0.001 in 100K steps. This extreme decrease suggests it is effectively a deterministic policy. \pke\ is achieved in a few hundred steps, usually before other effects become prominent (see Figure \ref{fig:reinforce_3best}), and is stronger than for Constant Reward.

	\subsection{NMT Experiments}\label{sec:peakiness_nmt}
	
	We turn to analyzing a real-world application of {\sc Reinforce} to NMT.
	Important differences between this and the previous simulations are: (1) it is rare in NMT for {\sc Reinforce} to sample from the same conditional distribution more than a handful of times, given the number of source sentences $x$ and sentence prefixes $y_{<i}$ (contexts); and (2) in NMT $P_\theta(\cdot|x,y_{<i})$ shares parameters between contexts,
	which means that updating $P_\theta$ for one context may influence $P_\theta$ for another. 
	
	We follow the same pretraining as in \S\ref{sec:peakiness_simulations}. We then follow \citet{yang2018improving} in defining the reward function based on the expected BLEU score. 
	Expected BLEU is computed by sampling suffixes for the sentence, and averaging the BLEU score of the sampled sentences against the reference.
	
	
	We use early stopping with a patience of 10 epochs, where each epoch consists of 5000 sentences sampled from the WMT2015 \cite{bojar2015Findings} German-English training data. We use $k=1$.
	We retuned the learning-rate, and positive baseline settings against
	the development set. Other hyper-parameters were an exact replication of the experiments reported in \citep{yang2018improving}.

	\paragraph{Results.} Results indicate an increase in the peakiness of the conditional distributions. Our results are based on a sample of 1000 contexts from the pretrained model, and another (independent) sample from the reinforced model.
	
\begin{figure}[t]
		\includegraphics[width=\columnwidth]{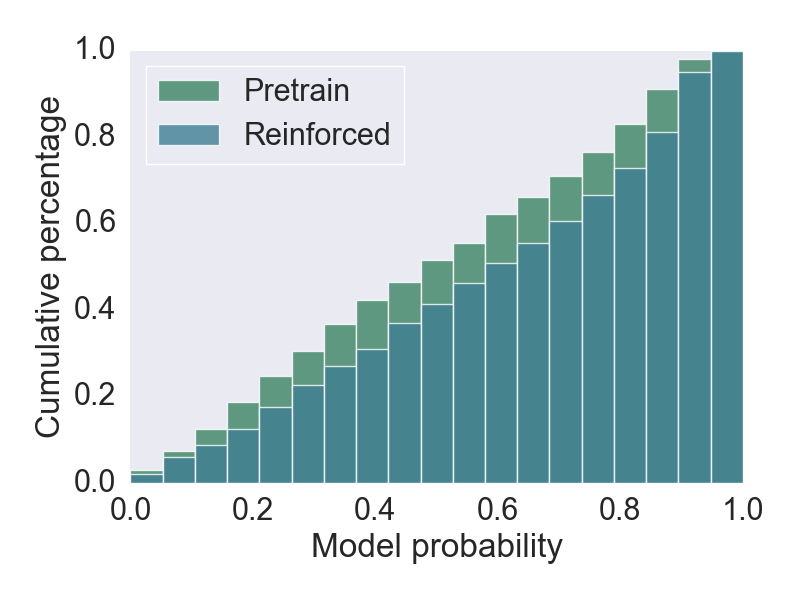}
		\caption{The cumulative distribution of the probability of the most likely token in the NMT experiments.
			The green distribution corresponds to the pretrained model,
			and the blue corresponds to the reinforced model.
			The y-axis is the proportion of conditional probabilities with a mode of value $\leq x$ (the x-axis). Note that a lower cumulative percentage means a more peaked output distribution. \label{fig:cum_tops}}
		\vspace{-0.4cm}
	\end{figure}

	Indeed, the modes of the conditional distributions tend to increase. Figure \ref{fig:cum_tops} presents the distribution
	of the modes' probability in the reinforced conditional distributions compared with the pretrained model, showing a shift of probability mass towards higher probabilities for the mode, following RL.
	Another indication of the increased peakiness is the decrease in the average entropy of $P_\theta$,
	which was reduced from 3.45 in the pretrained model to an average of 2.82 following RL. 
	This more modest reduction in entropy (compared to \S\ref{sec:peakiness_simulations}) might also suggest that the procedure did not converge
	to the optimal value for $\theta$, as then we would have expected the entropy to substantially drop if not to 0 (overfit), then to the average 
	entropy of valid next tokens (given the source and a prefix of the sentence).
	
	
	\vspace{-0.2cm}
	\section{Performance following {\sc Reinforce}}\label{sec:convergence_rate}
	
	We now turn to assessing under what conditions it is likely that {\sc Reinforce} will lead to an improvement in the performance
        of an NMT system. As in the previous section, we use both controlled simulations and NMT experiments.
        
	\subsection{Controlled Simulations}\label{sec:convergence_simulations}
	
	We use the same model and experimental setup described in Section \ref{sec:peakiness_simulations},
        this time only exploring the Simulated Reward setting,
	as a Constant Reward is not expected to converge to any meaningful $\theta$. Results are averaged over 100 conditional distributions
	sampled from the pretrained model.
	
	Caution should be exercised when determining the learning rate (LR).
	Common LRs used in the NMT literature are of the scale of $10^{-4}$. However, in our simulations, no LR smaller than 0.1 yielded any improvement in $R$. We thus set the LR to be 0.1.
	We note that in our simulations, a higher learning rate means faster convergence as our reward is noise-free: it is always highest for the best option. In practice, increasing
	the learning rate may deteriorate results, as it may cause the system to overfit to the sampled instances.
	Indeed, when increasing the learning rate in our NMT experiments (see below) by an order of magnitude, early stopping
	caused the RL procedure to stop without any parameter updates.
	
	Figure \ref{fig:reinforce_3best} shows how $P_\theta$ changes over the first 50K steps of {\sc Reinforce} (probabilities are averaged over 100 repetitions), 
	for a case where $y_{best}$ was initially the second, third and fourth most probable. 
	Although these are the easiest settings for {\sc Reinforce}, and despite the high learning rate, it fails to make $y_{best}$ the mode of the distribution within 100K steps, unless $y_{best}$ was initially the second most probable. In cases where $y_{best}$ is initially of a lower rank than four, it is hard to see any increase in its probability, even after 1M steps.
	
	\begin{figure}
		\includegraphics[width=\columnwidth]{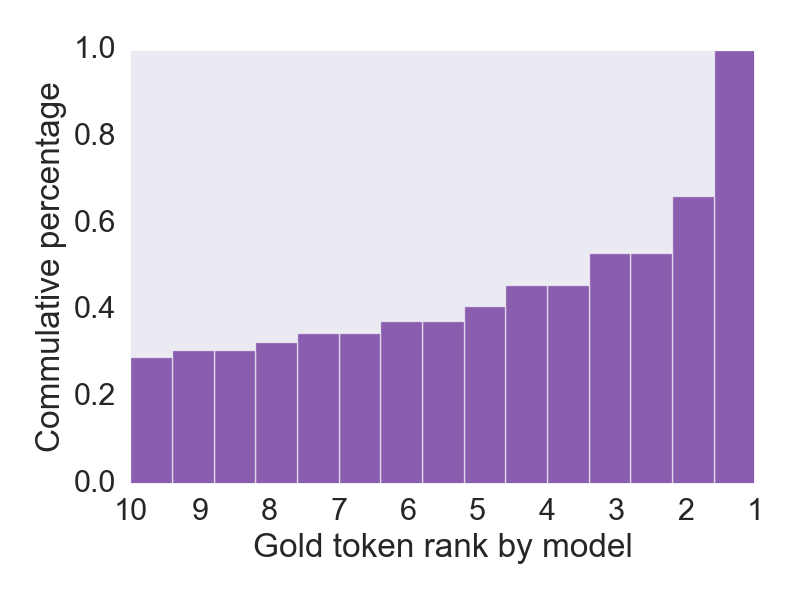}
		\caption{Cumulative percentage of contexts where the pretrained model ranks $y_{best}$ in rank $x$ or below and where it does not rank $y_{best}$ first ($x=0$).
			In about half the cases it is ranked fourth or below. \label{fig:ranks_till10}}
		\vspace{-0.4cm}
	\end{figure}
	
	\subsection{NMT Experiments}\label{sec:convergence_nmt_exps}
	
	We trained an NMT system, using the same procedure as in Section \ref{sec:peakiness_nmt}, and report BLEU scores over the news2014 test set. 
	After training with an expected BLEU reward, we indeed see a minor improvement which is consistent between trials 
	and pretrained models. While the pretrain BLEU score is 30.31, the reinforced one is 30.73. 
	
	Analyzing what words were influenced by the RL procedure, we begin by computing
	the cumulative probability of the target token $y_{best}$ to be ranked lower than a given rank according to the pretrained model.
	Results (Figure \ref{fig:ranks_till10}) show that in about half of the cases, $y_{best}$ is not among the top three choices of the pretrained model, and we thus expect it
	not to gain substantial probability following {\sc Reinforce}, according to our simulations.
	
\begin{figure}
		\includegraphics[width=\columnwidth]{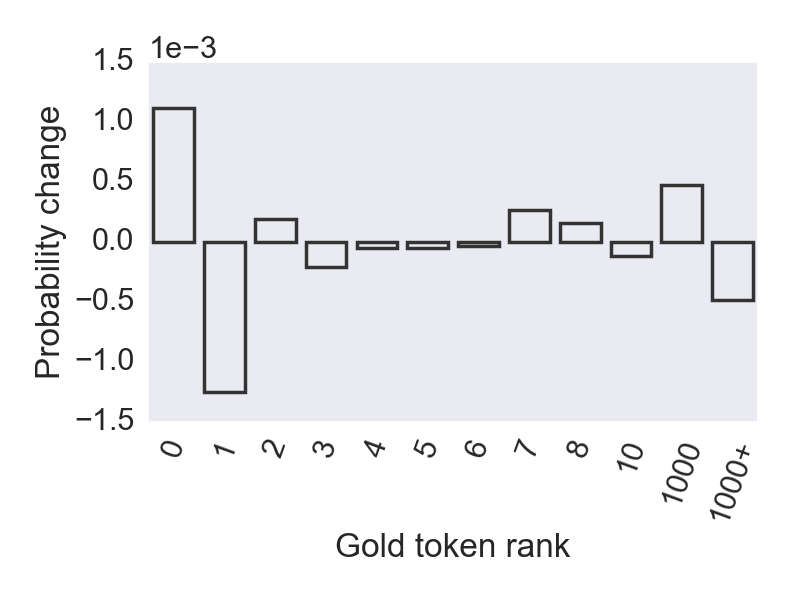}
		\caption{Difference between the ranks of $y_{best}$ in the reinforced and the pretrained model.
			Each column $x$ corresponds to the difference in the probability that $y_{best}$ is ranked in rank $x$ in the reinforced 
			model, and the same probability in the pretrained model.\label{fig:bleu_rank_diffs}}
		\vspace{-0.4cm}
	\end{figure}

	We next turn to compare the ranks the reinforced model assigns to the target tokens, and their ranks according to the pretrained model.
	Figure \ref{fig:bleu_rank_diffs} presents the difference in the probability that $y_{best}$ 
	is ranked at a given rank following RL and the probability it is ranked there initially.
	Results indicate that indeed more
	target tokens are ranked first, and less second, but little consistent shift of probability mass occurs otherwise across the ten first ranks. 
	It is possible that RL has managed to push $y_{best}$ in some cases between very low ranks (<1000) to medium-low ranks (between 10 and 1000).
	However, token probabilities  in these ranks are so low that it is unlikely to affect the system outputs in any way.
	This fits well with the results of our simulations that predicted that only the initially top-ranked tokens are likely to change.
	
	In an attempt to explain the improved BLEU score following RL with \pke, we repeat the NMT experiment this time using
	a constant reward of 1. 
	Our results present a nearly identical improvement in BLEU, achieving 30.72, and a similar pattern in the change of the target tokens' ranks (see Appendix \ref{fig:constant_rank_diffs}).
	Therefore, there is room to suspect that even in cases where RL yields an improvement in BLEU, it may 
	partially result from reward-independent factors, such as \pke.\footnote{We tried several other reward functions as well, 
		all of which got BLEU scores of 30.73--30.84. This improvement is very stable across metrics, trials and pretrained models.}
	
	\section{Experiments with Contrastive MRT}\label{sec:MRT_exp}
	
	\begin{figure*}
		\begin{subfigure}[b]{.5\textwidth}
			\includegraphics[width=\columnwidth]{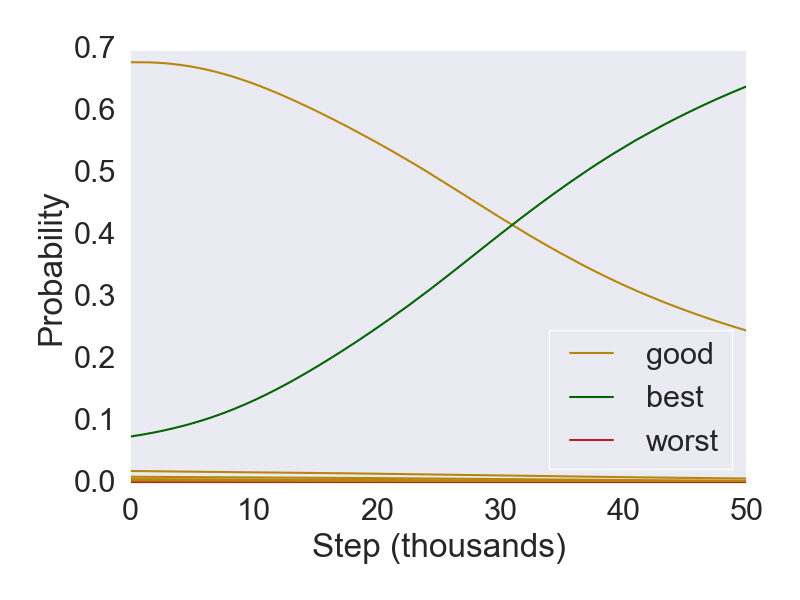}
		\end{subfigure}
		\begin{subfigure}[b]{.5\textwidth}
			\includegraphics[width=\columnwidth]{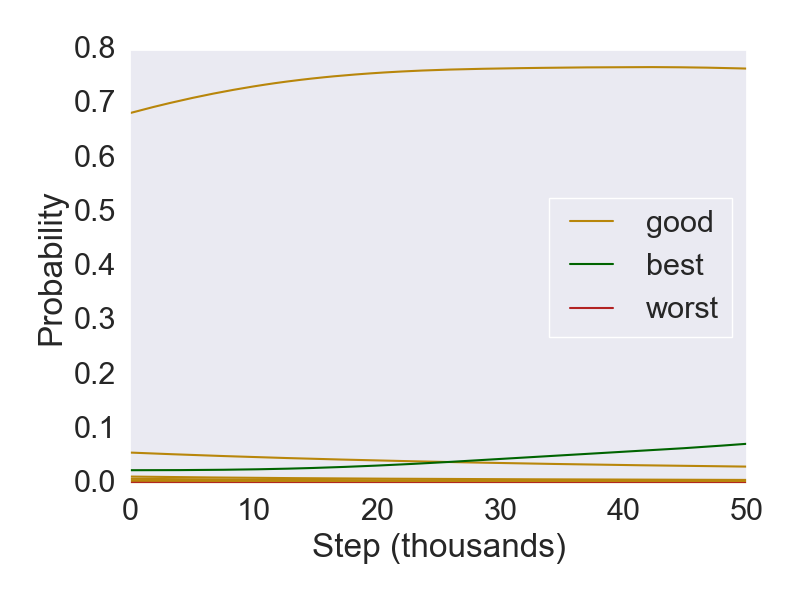}
		\end{subfigure}
		\caption{
			The probability of different tokens following CMRT, in the controlled simulations in the Simulated Reward setting.
			The left/right figures correspond to simulations where the target token ($y_{best}$) was initially the second/third
			most probable token. The green line corresponds to the target token, yellow lines to medium-reward tokens and red lines to tokens with $r(y)=0$.
			\label{fig:mrt}}
		\vspace{-0.4cm}
	\end{figure*}

	In \S\ref{sec:mrt_theory} we showed that CMRT does not, in fact, maximize $R$, and so does not enjoy the same theoretical guarantees
	as {\sc Reinforce} and similar policy gradient methods. However, it is still the RL procedure of choice in much recent work on NMT.
	We therefore repeat the simulations described in \S\ref{sec:pke} and \S\ref{sec:convergence_rate}, assessing the performance of MRT in these conditions. 
	We experiment with $\alpha=0.005$ and $k=20$, common settings in the literature, and average over 100 trials.
	
	Figure \ref{fig:mrt} shows how the distribution $P_\theta$ changes over the course of 50K update steps to $\theta$, where $y_{best}$
	is taken to be the second and third initially most probable token (Simulated Reward setting).
	Results are similar in trends to those obtained with {\sc Reinforce}: MRT succeeds in pushing $y_{best}$ to be the highest ranked token if it was initially second,
	but struggles where it was initially ranked third or below.
	We only observe a small \pke\ in MRT. This is probably due to the contrastive effect, which means that tokens that were not sampled 
	do not lose probability mass.
	
	All graphs we present here allow sampling the same token more than once in each batch (i.e., $S$ is a sample with replacements).
	Simulations with deduplication show similar results.

	\vspace{-.1cm}
	\section{Discussion}\label{sec:discussion}
	\vspace{-.1cm}
	
	
	In this paper, we showed that the type of distributions used in NMT entail that promoting the target token to be the mode is likely to take a prohibitively long times for existing RL practices, except under the best conditions (where the pretrained model is ``nearly'' correct). 
	This leads us to conclude that observed improvements from using RL for NMT are likely due either to fine-tuning
	the most probable tokens in the pretrained model (an effect which may be more easily achieved using reranking methods, and uses but little of the power of RL methods),
        or to effects unrelated to the signal carried by the reward, such as \pke.
	Another contribution of this paper is in showing that CMRT does not optimize the expected reward and is thus theoretically unmotivated.
	
	A number of reasons lead us to believe that in our NMT experiments, improvements are not due to the reward function, but to
	artefacts such as \pke. First, reducing a constant baseline from $r$, so as to make the expected reward zero, disallows learning.
	This is surprising, as {\sc Reinforce}, generally and in our simulations, converges faster where the reward is centered around zero,
	and so the fact that this procedure here disallows learning hints that other factors are in play. 
	As \pke\ can be observed even where the reward is constant (if the expected reward is positive; see \S\ref{sec:peakiness_simulations}),
	this suggests \pke\ may play a role here.
	Second, we observe more peakiness in the reinforced model and in such cases, we expect improvements in BLEU \cite{caccia2018language}.
	Third, we achieve similar results with a constant reward in our NMT experiments (\S\ref{sec:convergence_nmt_exps}).
	Fourth, our controlled simulations show that asymptotic convergence is not reached in any but the
        easiest conditions (\S\ref{sec:convergence_simulations}).
	
	Our analysis further suggests that gradient clipping, sometimes used in NMT \citep{zhang2016generating},
	is expected to hinder convergence further. It should be avoided when using {\sc Reinforce} as it violates {\sc Reinforce}'s assumptions. 

	The per-token sampling as done in our experiments is more exploratory than beam search \cite{Wu2018ASO}, reducing \pke. 
     Furthermore, the latter does not sample from the behavior policy, but does not properly account for being off-policy in the parameter updates.
	
	Adding the reference to the sample $S$, which some implementations allow \citep{sennrich2017nematus}
        may help reduce the problems of never sampling the target tokens.
	However, as \citet{edunov2018classical} point out, this practice may lower results, as 
	it may destabilize training by leading the model to improve over outputs it cannot generalize over, as they are very different from
	anything the model assigns a high probability to, at the cost of other outputs.
	
	\vspace{-.1cm}
        \section{Conclusion}
	\vspace{-.1cm}

	The standard MT scenario poses several uncommon challenges for RL. First, the action space in MT problems is a high-dimensional discrete space (generally in the size of the vocabulary of the target language or the product thereof for sentences). This contrasts with the more common scenario studied by contemporary RL methods, which focuses mostly on much smaller discrete action spaces (e.g., video games \citep{mnih2015human,mnih2016asynchronous}), or continuous action spaces of relatively low dimensions (e.g., simulation of robotic control tasks \citep{lillicrap2015continuous}). Second, reward for MT is naturally very sparse -- almost all possible sentences are ``wrong'' (hence, not rewarding) in a given context. Finally, it is common in MT to use RL for tuning a pretrained model. Using a pretrained model ameliorates the last problem. But then, these pretrained models are in general quite peaky, and because training is done \textit{on-policy} -- that is, actions are being sampled from the same model being optimized -- exploration is inherently limited.
	
	Here we argued that, taken together, these challenges result in significant weaknesses for current RL practices for NMT, 
	that may ultimately prevent them from being truly useful. 
	At least some of these challenges have been widely studied in the RL literature, with numerous techniques developed to address them,
	but were not yet adopted in NLP. We turn to discuss some of them.
	
	Off-policy methods, in which observations are sampled from a different policy than the one being currently optimized, are prominent in RL \citep{watkins1992q, sutton1998reinforcement}, and were also studied in the context of policy gradient methods \citep{degris2012off, silver2014deterministic}. In principle, such methods allow learning from a more ``exploratory'' policy. Moreover, a key motivation for using $\alpha$ in CMRT is smoothing; off-policy sampling allows smoothing while keeping convergence guarantees.
	
	In its basic form, exploration in {\sc Reinforce} relies on stochasticity in the action-selection (in MT, this is due to sampling). More sophisticated exploration methods have been extensively studied, for example using measures for the exploratory usefulness of states or actions \citep{fox2018dora}, or relying on parameter-space noise rather than action-space noise \citep{plappert2017parameter}.
	
	For MT, an additional challenge is that even effective exploration (sampling diverse sets of observations), may not be enough, since the state-action space is too large to be effectively covered, with almost all sentences being not rewarding. Recently, diversity-based and multi-goal methods for RL were proposed to tackle similar challenges \citep{andrychowicz2017hindsight,ghosh2018divideandconquer,eysenbach2018diversity}. 
	We believe the adoption of such methods is a promising path forward for the application of RL in NLP.


	%
	
	%
	%
	
	\section{Acknowledgments}
	This work was supported by the Israel Science Foundation (grant no. 929/17) and by the HUJI Cyber Security Research Center in conjunction with the Israel National Cyber Bureau in the Prime Minister's Office.
	
	\bibliography{flaws_rl}

\begin{thebibliography}{46}
\expandafter\ifx\csname natexlab\endcsname\relax\def\natexlab#1{#1}\fi

\bibitem[{Andrychowicz et~al.(2017)Andrychowicz, Wolski, Ray, Schneider, Fong,
  Welinder, McGrew, Tobin, Abbeel, and Zaremba}]{andrychowicz2017hindsight}
Marcin Andrychowicz, Filip Wolski, Alex Ray, Jonas Schneider, Rachel Fong,
  Peter Welinder, Bob McGrew, Josh Tobin, OpenAI~Pieter Abbeel, and Wojciech
  Zaremba. 2017.
\newblock Hindsight experience replay.
\newblock In \emph{Advances in Neural Information Processing Systems}, pages
  5048--5058.

\bibitem[{Ayana et~al.(2016)Ayana, Liu, and Sun}]{ayana2016neural}
Shiqi~Shen Ayana, Zhiyuan Liu, and Maosong Sun. 2016.
\newblock Neural headline generation with minimum risk training.
\newblock \emph{arXiv preprint arXiv:1604.01904}.

\bibitem[{Bojar et~al.(2015)Bojar, Chatterjee, Federmann, Haddow, Huck, Hokamp,
  Koehn, Logacheva, Monz, Negri, Post, Scarton, Specia, and
  Turchi}]{bojar2015Findings}
Ondrej Bojar, Rajen Chatterjee, Christian Federmann, Barry Haddow, Matthias
  Huck, Chris Hokamp, Philipp Koehn, Varvara Logacheva, Christof Monz, Matteo
  Negri, Matt Post, Carolina Scarton, Lucia Specia, and Marco Turchi. 2015.
\newblock Findings of the 2015 workshop on statistical machine translation.
\newblock In \emph{WMT@EMNLP}.

\bibitem[{Caccia et~al.(2018)Caccia, Caccia, Fedus, Larochelle, Pineau, and
  Charlin}]{caccia2018language}
Massimo Caccia, Lucas Caccia, William Fedus, Hugo Larochelle, Joelle Pineau,
  and Laurent Charlin. 2018.
\newblock \href {https://arxiv.org/pdf/1811.02549.pdf} {Language gans falling
  short}.
\newblock \emph{arXiv preprint arXiv:1811.02549}.

\bibitem[{Degris et~al.(2012)Degris, White, and Sutton}]{degris2012off}
Thomas Degris, Martha White, and Richard~S Sutton. 2012.
\newblock Off-policy actor-critic.
\newblock \emph{arXiv preprint arXiv:1205.4839}.

\bibitem[{Edunov et~al.(2018)Edunov, Ott, Auli, Grangier, and
  Ranzato}]{edunov2018classical}
Sergey Edunov, Myle Ott, Michael Auli, David Grangier, and Marc'Aurelio
  Ranzato. 2018.
\newblock \href {https://doi.org/10.18653/v1/N18-1033} {Classical structured
  prediction losses for sequence to sequence learning}.
\newblock In \emph{Proceedings of the 2018 Conference of the North American
  Chapter of the Association for Computational Linguistics: Human Language
  Technologies, Volume 1 (Long Papers)}, pages 355--364. Association for
  Computational Linguistics.

\bibitem[{Eysenbach et~al.(2019)Eysenbach, Gupta, Ibarz, and
  Levine}]{eysenbach2018diversity}
Benjamin Eysenbach, Abhishek Gupta, Julian Ibarz, and Sergey Levine. 2019.
\newblock \href {https://openreview.net/forum?id=SJx63jRqFm} {Diversity is all
  you need: Learning skills without a reward function}.
\newblock In \emph{International Conference on Learning Representations}.

\bibitem[{Fox et~al.(2018)Fox, Choshen, and Loewenstein}]{fox2018dora}
Lior Fox, Leshem Choshen, and Yonatan Loewenstein. 2018.
\newblock Dora the explorer: Directed outreaching reinforcement
  action-selection.
\newblock \emph{ICLR}, abs/1804.04012.

\bibitem[{Ghosh et~al.(2018)Ghosh, Singh, Rajeswaran, Kumar, and
  Levine}]{ghosh2018divideandconquer}
Dibya Ghosh, Avi Singh, Aravind Rajeswaran, Vikash Kumar, and Sergey Levine.
  2018.
\newblock \href {https://openreview.net/forum?id=rJwelMbR-} {Divide-and-conquer
  reinforcement learning}.
\newblock In \emph{International Conference on Learning Representations}.

\bibitem[{Hendricks et~al.(2016)Hendricks, Akata, Rohrbach, Donahue, Schiele,
  and Darrell}]{Hendricks2016Generating}
Lisa~Anne Hendricks, Zeynep Akata, Marcus Rohrbach, Jeff Donahue, Bernt
  Schiele, and Trevor Darrell. 2016.
\newblock Generating visual explanations.
\newblock In \emph{ECCV}.

\bibitem[{Kingma and Ba(2015)}]{kingma2015Adam}
Diederik~P. Kingma and Jimmy Ba. 2015.
\newblock Adam: A method for stochastic optimization.
\newblock \emph{CoRR}, abs/1412.6980.

\bibitem[{Koehn et~al.(2007)Koehn, Hoang, Birch, Callison-Burch, Federico,
  Bertoldi, Cowan, Shen, Moran, Zens, Dyer, Bojar, Constantin, and
  Herbst}]{koehn2007koehn}
Philipp Koehn, Hieu Hoang, Alexandra Birch, Chris Callison-Burch, Marcello
  Federico, Nicola Bertoldi, Brooke Cowan, Wade Shen, Christine Moran, Richard
  Zens, Chris Dyer, Ondrej Bojar, Alexandra Constantin, and Evan Herbst. 2007.
\newblock Moses: Open source toolkit for statistical machine translation.
\newblock In \emph{Proceedings of the 45th Annual Meeting of the Association
  for Computational Linguistics Companion Volume Proceedings of the Demo and
  Poster Sessions}, pages 177--180.

\bibitem[{Li et~al.(2017)Li, Monroe, Shi, Jean, Ritter, and
  Jurafsky}]{li2017adversarial}
Jiwei Li, Will Monroe, Tianlin Shi, S\'ebastien Jean, Alan Ritter, and Dan
  Jurafsky. 2017.
\newblock \href {https://www.aclweb.org/anthology/D17-1230} {Adversarial
  learning for neural dialogue generation}.
\newblock In \emph{Proceedings of the 2017 Conference on Empirical Methods in
  Natural Language Processing}, pages 2157--2169, Copenhagen, Denmark.
  Association for Computational Linguistics.

\bibitem[{Lillicrap et~al.(2015)Lillicrap, Hunt, Pritzel, Heess, Erez, Tassa,
  Silver, and Wierstra}]{lillicrap2015continuous}
Timothy~P Lillicrap, Jonathan~J Hunt, Alexander Pritzel, Nicolas Heess, Tom
  Erez, Yuval Tassa, David Silver, and Daan Wierstra. 2015.
\newblock Continuous control with deep reinforcement learning.
\newblock \emph{arXiv preprint arXiv:1509.02971}.

\bibitem[{Liu et~al.(2016)Liu, Zhu, Ye, Guadarrama, and
  Murphy}]{liu2016optimization}
Siqi Liu, Zhenhai Zhu, Ning Ye, Sergio Guadarrama, and Kevin Murphy. 2016.
\newblock Optimization of image description metrics using policy gradient
  methods.
\newblock \emph{CoRR, abs/1612.00370}, 2.

\bibitem[{Makarov and Clematide(2018)}]{makarov2018Neural}
Peter Makarov and Simon Clematide. 2018.
\newblock Neural transition-based string transduction for limited-resource
  setting in morphology.
\newblock In \emph{COLING}.

\bibitem[{Mnih et~al.(2016)Mnih, Badia, Mirza, Graves, Lillicrap, Harley,
  Silver, and Kavukcuoglu}]{mnih2016asynchronous}
Volodymyr Mnih, Adria~Puigdomenech Badia, Mehdi Mirza, Alex Graves, Timothy
  Lillicrap, Tim Harley, David Silver, and Koray Kavukcuoglu. 2016.
\newblock Asynchronous methods for deep reinforcement learning.
\newblock In \emph{International conference on machine learning}, pages
  1928--1937.

\bibitem[{Mnih et~al.(2015)Mnih, Kavukcuoglu, Silver, Rusu, Veness, Bellemare,
  Graves, Riedmiller, Fidjeland, Ostrovski et~al.}]{mnih2015human}
Volodymyr Mnih, Koray Kavukcuoglu, David Silver, Andrei~A Rusu, Joel Veness,
  Marc~G Bellemare, Alex Graves, Martin Riedmiller, Andreas~K Fidjeland, Georg
  Ostrovski, et~al. 2015.
\newblock Human-level control through deep reinforcement learning.
\newblock \emph{Nature}, 518(7540):529--533.

\bibitem[{Neubig(2016)}]{neubig2016Lexicons}
Graham Neubig. 2016.
\newblock Lexicons and minimum risk training for neural machine translation:
  Naist-cmu at wat2016.
\newblock In \emph{WAT@COLING}.

\bibitem[{Neubig et~al.(2018)Neubig, Sperber, Wang, Felix, Matthews,
  Padmanabhan, Qi, Sachan, Arthur, Godard, Hewitt, Riad, and
  Wang}]{neubig2018XNMTTE}
Graham Neubig, Matthias Sperber, Xinyi Wang, Matthieu Felix, Austin Matthews,
  Sarguna Padmanabhan, Ye~Qi, Devendra~Singh Sachan, Philip Arthur, Pierre
  Godard, John Hewitt, Rachid Riad, and Liming Wang. 2018.
\newblock Xnmt: The extensible neural machine translation toolkit.
\newblock In \emph{AMTA}.

\bibitem[{Och(2003)}]{och2003minimum}
Franz~Josef Och. 2003.
\newblock Minimum error rate training in statistical machine translation.
\newblock In \emph{Proceedings of the 41st Annual Meeting on Association for
  Computational Linguistics-Volume 1}, pages 160--167. Association for
  Computational Linguistics.

\bibitem[{Papineni et~al.(2002)Papineni, Roukos, Ward, and
  Zhu}]{papineni2002bleu}
Kishore Papineni, Salim Roukos, Todd Ward, and Wei-Jing Zhu. 2002.
\newblock \href {https://www.aclweb.org/anthology/P02-1040.pdf} {Bleu: a method
  for automatic evaluation of machine translation}.
\newblock In \emph{Proceedings of the 40th annual meeting on association for
  computational linguistics}, pages 311--318. Association for Computational
  Linguistics.

\bibitem[{Plappert et~al.(2017)Plappert, Houthooft, Dhariwal, Sidor, Chen,
  Chen, Asfour, Abbeel, and Andrychowicz}]{plappert2017parameter}
Matthias Plappert, Rein Houthooft, Prafulla Dhariwal, Szymon Sidor, Richard~Y
  Chen, Xi~Chen, Tamim Asfour, Pieter Abbeel, and Marcin Andrychowicz. 2017.
\newblock Parameter space noise for exploration.
\newblock \emph{arXiv preprint arXiv:1706.01905}.

\bibitem[{Press et~al.(2017)Press, Bar, Bogin, Berant, and
  Wolf}]{press2017language}
O.~Press, A.~Bar, B.~Bogin, J.~Berant, and L.~Wolf. 2017.
\newblock Language generation with recurrent generative adversarial networks
  without pre-training.
\newblock In \emph{Fist Workshop on Learning to Generate Natural
  Language@ICML}.

\bibitem[{Ranzato et~al.(2015)Ranzato, Chopra, Auli, and
  Zaremba}]{ranzato2015sequence}
Marc'Aurelio Ranzato, Sumit Chopra, Michael Auli, and Wojciech Zaremba. 2015.
\newblock Sequence level training with recurrent neural networks.
\newblock \emph{arXiv preprint arXiv:1511.06732}.

\bibitem[{Rennie et~al.(2017)Rennie, Marcheret, Mroueh, Ross, and
  Goel}]{rennie2017self}
Steven~J Rennie, Etienne Marcheret, Youssef Mroueh, Jerret Ross, and Vaibhava
  Goel. 2017.
\newblock Self-critical sequence training for image captioning.
\newblock In \emph{Proceedings of the IEEE Conference on Computer Vision and
  Pattern Recognition}, pages 7008--7024.

\bibitem[{Sakaguchi et~al.(2017)Sakaguchi, Post, and
  Van~Durme}]{sakaguchi2017grammatical}
Keisuke Sakaguchi, Matt Post, and Benjamin Van~Durme. 2017.
\newblock Grammatical error correction with neural reinforcement learning.
\newblock \emph{arXiv preprint arXiv:1707.00299}.

\bibitem[{Schulz et~al.(2018)Schulz, Aziz, and Cohn}]{Schulz2018ASD}
Philip Schulz, Wilker Aziz, and Trevor Cohn. 2018.
\newblock A stochastic decoder for neural machine translation.
\newblock In \emph{ACL}.

\bibitem[{Sennrich et~al.(2017)Sennrich, Firat, Cho, Birch, Haddow, Hitschler,
  Junczys-Dowmunt, L{\"a}ubli, Barone, Mokry, and
  Nadejde}]{sennrich2017nematus}
Rico Sennrich, Orhan Firat, Kyunghyun Cho, Alexandra Birch, Barry Haddow,
  Julian Hitschler, Marcin Junczys-Dowmunt, Samuel L{\"a}ubli, Antonio
  Valerio~Miceli Barone, Jozef Mokry, and Maria Nadejde. 2017.
\newblock Nematus: a toolkit for neural machine translation.
\newblock In \emph{EACL}.

\bibitem[{Sennrich et~al.(2016)Sennrich, Haddow, and
  Birch}]{sennrich2016neural}
Rico Sennrich, Barry Haddow, and Alexandra Birch. 2016.
\newblock \href {http://www.aclweb.org/anthology/P16-1162} {Neural machine
  translation of rare words with subword units}.
\newblock In \emph{Proceedings of the 54th Annual Meeting of the Association
  for Computational Linguistics (Volume 1: Long Papers)}, volume~1, pages
  1715--1725.

\bibitem[{Shen et~al.(2016)Shen, Cheng, He, He, Wu, Sun, and
  Liu}]{shen2016minimum}
Shiqi Shen, Yong Cheng, Zhongjun He, Wei He, Hua Wu, Maosong Sun, and Yang Liu.
  2016.
\newblock \href {https://doi.org/10.18653/v1/P16-1159} {Minimum risk training
  for neural machine translation}.
\newblock In \emph{Proceedings of the 54th Annual Meeting of the Association
  for Computational Linguistics (Volume 1: Long Papers)}, pages 1683--1692.
  Association for Computational Linguistics.

\bibitem[{Shen et~al.(2017)Shen, Liu, and Sun}]{shen2017Optimizing}
Shiqi Shen, Yang Liu, and Maosong Sun. 2017.
\newblock Optimizing non-decomposable evaluation metrics for neural machine
  translation.
\newblock \emph{Journal of Computer Science and Technology}, 32:796--804.

\bibitem[{Shetty et~al.(2017)Shetty, Rohrbach, Hendricks, Fritz, and
  Schiele}]{shetty2017speaking}
Rakshith Shetty, Marcus Rohrbach, Lisa~Anne Hendricks, Mario Fritz, and Bernt
  Schiele. 2017.
\newblock Speaking the same language: Matching machine to human captions by
  adversarial training.
\newblock In \emph{2017 IEEE International Conference on Computer Vision
  (ICCV)}, pages 4155--4164. IEEE.

\bibitem[{Silver et~al.(2014)Silver, Lever, Heess, Degris, Wierstra, and
  Riedmiller}]{silver2014deterministic}
David Silver, Guy Lever, Nicolas Heess, Thomas Degris, Daan Wierstra, and
  Martin Riedmiller. 2014.
\newblock Deterministic policy gradient algorithms.
\newblock In \emph{ICML}.

\bibitem[{Sutton and Barto(1998)}]{sutton1998reinforcement}
Richard~S Sutton and Andrew~G Barto. 1998.
\newblock \emph{Reinforcement learning: An introduction}.
\newblock MIT press.

\bibitem[{Tevet et~al.(2018)Tevet, Habib, Shwartz, and
  Berant}]{tevet2018evaluating}
G.~Tevet, G.~Habib, V.~Shwartz, and J.~Berant. 2018.
\newblock Evaluating text {GAN}s as language models.
\newblock \emph{arXiv preprint arXiv:1810.12686}.

\bibitem[{Tieleman and Hinton(2012)}]{tieleman2012lecture}
Tijmen Tieleman and Geoffrey Hinton. 2012.
\newblock Lecture 6.5-rmsprop: Divide the gradient by a running average of its
  recent magnitude.
\newblock \emph{COURSERA: Neural networks for machine learning}, 4(2):26--31.

\bibitem[{Vaswani et~al.(2017)Vaswani, Shazeer, Parmar, Uszkoreit, Jones,
  Gomez, Kaiser, and Polosukhin}]{vaswani2017attention}
Ashish Vaswani, Noam Shazeer, Niki Parmar, Jakob Uszkoreit, Llion Jones,
  Aidan~N Gomez, {\L}ukasz Kaiser, and Illia Polosukhin. 2017.
\newblock \href
  {https://papers.nips.cc/paper/7181-attention-is-all-you-need.pdf} {Attention
  is all you need}.
\newblock In \emph{Advances in Neural Information Processing Systems}, pages
  5998--6008.

\bibitem[{Watkins and Dayan(1992)}]{watkins1992q}
Christopher~JCH Watkins and Peter Dayan. 1992.
\newblock Q-learning.
\newblock \emph{Machine learning}, 8(3-4):279--292.

\bibitem[{Williams(1992)}]{williams1992simple}
Ronald~J Williams. 1992.
\newblock Simple statistical gradient-following algorithms for connectionist
  reinforcement learning.
\newblock \emph{Machine learning}, 8(3-4):229--256.

\bibitem[{Wu et~al.(2018)Wu, Tian, Qin, Lai, and Liu}]{Wu2018ASO}
Lijun Wu, Fei Tian, Tao Qin, Jianhuang Lai, and Tie-Yan Liu. 2018.
\newblock A study of reinforcement learning for neural machine translation.
\newblock In \emph{EMNLP}.

\bibitem[{Wu et~al.(2017)Wu, Xia, Zhao, Tian, Qin, Lai, and
  Liu}]{wu2017adversarial}
Lijun Wu, Yingce Xia, Li~Zhao, Fei Tian, Tao Qin, Jianhuang Lai, and Tie-Yan
  Liu. 2017.
\newblock Adversarial neural machine translation.
\newblock \emph{arXiv preprint arXiv:1704.06933}.

\bibitem[{Yang et~al.(2018)Yang, Chen, Wang, and Xu}]{yang2018improving}
Zhen Yang, Wei Chen, Feng Wang, and Bo~Xu. 2018.
\newblock \href {https://doi.org/10.18653/v1/N18-1122} {Improving neural
  machine translation with conditional sequence generative adversarial nets}.
\newblock In \emph{Proceedings of the 2018 Conference of the North American
  Chapter of the Association for Computational Linguistics: Human Language
  Technologies, Volume 1 (Long Papers)}, pages 1346--1355. Association for
  Computational Linguistics.

\bibitem[{Yu et~al.(2017)Yu, Zhang, Wang, and Yu}]{yu2017seqgan}
Lantao Yu, Weinan Zhang, Jun Wang, and Yong Yu. 2017.
\newblock Seqgan: Sequence generative adversarial nets with policy gradient.
\newblock In \emph{AAAI}, pages 2852--2858.

\bibitem[{Zhang et~al.(2017)Zhang, Ding, Shen, Cheng, Sun, Luan, and
  Liu}]{zhang2017thumt}
Jiac~heng Zhang, Yanzhuo Ding, Shiqi Shen, Yong Cheng, Maosong Sun, Huanbo
  Luan, and Yang Liu. 2017.
\newblock Thumt: An open source toolkit for neural machine translation.
\newblock \emph{arXiv preprint arXiv:1706.06415}.

\bibitem[{Zhang et~al.(2016)Zhang, Gan, and Carin}]{zhang2016generating}
Yizhe Zhang, Zhe Gan, and Lawrence Carin. 2016.
\newblock Generating text via adversarial training.
\newblock In \emph{NIPS workshop on Adversarial Training}, volume~21.

\end{thebibliography}
	\bibliographystyle{acl_natbib}
	
	\appendix
	\section{Contrastive MRT does not Maximize the Expected Reward}\label{app:counter_ex}
	
	We hereby detail a simple example where following the Contrastive MRT method (see \S\ref{sec:mrt_theory}) does not converge to the parameter value that maximizes $R$.
	
	Let $\theta$ be a real number in $[0,0.5]$, and let $P_{\theta}$ be a family of distributions over three values $a,b,c$ such that:
	
	$$
	P_{_\theta}(x)=
	\begin{cases}
	\theta & x=a\\
	2\theta^2 & x=b\\
	1-\theta-2\theta^2 & x=c\\
	\end{cases}
	$$
	
	Let $r(a)=1,r(b)=0,r(c)=0.5$.
	The expected reward as a function of $\theta$ is:
	
	$$R(\theta) = \theta + 0.5(1-\theta-2\theta^2)$$
	
	\noindent
	$R(\theta)$ is uniquely maximized by $\theta^*$ = 0.25. 
	
	Table \ref{tab:counter_ex} details the possible samples of size $k=2$,
	their probabilities, the corresponding $\widetilde{R}$ and its gradient.
	Standard numerical methods show that $\EX[\nabla\widetilde{R}]$ over possible
	samples $S$ is positive for $\theta \in (0,\gamma)$
	and negative for $\theta \in (\gamma,0.5]$, where $\gamma \approx 0.295$.
	This means that for any initialization of $\theta \in (0,0.5]$, Contrastive
	MRT will converge to $\gamma$ if the learning rate is sufficiently small. 
	For $\theta=0$, $\widetilde{R} \equiv 0.5$, and there will be no gradient updates, so 
	the method will converge to $\theta=0$. Neither
	of these values maximizes $R(\theta)$.
	
	We note that by using some $g\left(\theta\right)$ the $\gamma$ could be arbitrarily far from $\theta^*$.  $g$ could also map to $\left(-inf, inf\right)$ more often used in neural networks parameters.
	
	We further note that resorting to maximizing $\EX[\widetilde{R}]$ instead, does not maximize $R(\theta)$ either. Indeed, plotting $\EX[\widetilde{R}]$ 
	as a function of $\theta$ for this example, yields a maximum at $\theta \approx 0.32$.

	\begin{table}[h]
		\begin{tabular}{c|c|c|c|}
			$S$ & $P(S)$ & $\widetilde{R}$ & $\nabla\widetilde{R}$ \\
			\hline
			\hline
			$\{a,b\}$ & $4\theta^3$ & $\frac{1}{1+2\theta}$ & $\frac{-2}{(1+2\theta)^2}$ \\
			\hline
			$\{a,c\}$ & $2\theta(1$-$\theta$-$2\theta^2)$ & 0.5 + $\frac{\theta}{2-4\theta^2}$ &  $\frac{2 x^{2}+1}{2\left(1-2 \theta^{2}\right)^{2}} $\\
			\hline
			$\{b,c\}$ & $4\theta^2(1$-$\theta$-$2\theta^2)$  & 
			$\frac{1-\theta-2\theta^2}{2-2\theta}$ & 
			$\frac{\theta^2-2\theta}{(1-\theta)^{2}}$ \\
			\hline
			$a,a$ & $\theta^2$ & 1 & 0 \\ 
			\hline
			$b,b$ & $4\theta^4$ & 0 & 0 \\ 
			\hline
			$c,c$ & $(1$-$\theta$-$2\theta^2)^2$ & 0.5 & 0\\ 
			\hline
		\end{tabular}
		\caption{The gradients of $\widetilde{R}$ for each possible sample $S$. The batch size is $k=2$. Rows correspond to different sampled outcomes. $\nabla\widetilde{R}$ is the gradient of $\widetilde{R}$ given the corresponding value for $S$.\label{tab:counter_ex}}
	\end{table}

	\section{Deriving the Gradient of $\widetilde{R}$}\label{app:grad_mrt}
	
	Given $S$, recall the definition of $\widetilde{R}$:
	
	$$\widetilde{R}(\theta,S) = \sum_{i=1}^k Q_{\theta,S}(y_i)r(y_i) $$
	
	Taking the deriviative w.r.t. $\theta$:
	
	{\small
		$$\sum_{i=1}^k r(y_i)\frac{\nabla{}P(y)\cdot \alpha P(y)^{\alpha-1} \cdot Z(S) - \nabla{}Z(S) \cdot P(y)^\alpha}{Z(S)^2} = $$
		
		$$\sum_{i=1}^k r(y_i) \Big(\frac{\alpha\nabla{}P(y_i)}{P(y_i)}Q(y_i) - \frac{\nabla{}Z(S)}{Z(S)}Q(y_i)\Big) = $$
		
		$$\sum_{i=1}^k r(y_i) Q(y_i) \Big(\alpha\nabla{}\log{}P(y_i)-\nabla\log{}Z(S)\Big) = $$
		
		$$ \alpha \sum_{i=1}^k \Big( r(y_i)Q(y_i) \nabla\log~P(y_i) \Big) - \EX_{Q}[r]\nabla\log{}Z(S)$$
	}

	\begin{figure*}[th]
		\begin{subfigure}[b]{0.33\textwidth}
			\includegraphics[width=\columnwidth]{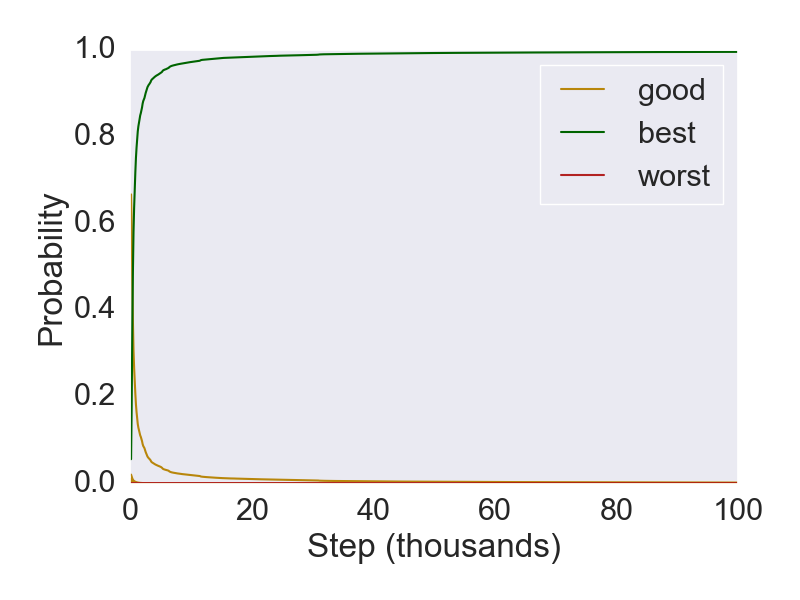}
			\caption{\label{fig:reinforce-1_best1}}
		\end{subfigure}
		\begin{subfigure}[b]{0.33\textwidth} 
			\includegraphics[width=\columnwidth]{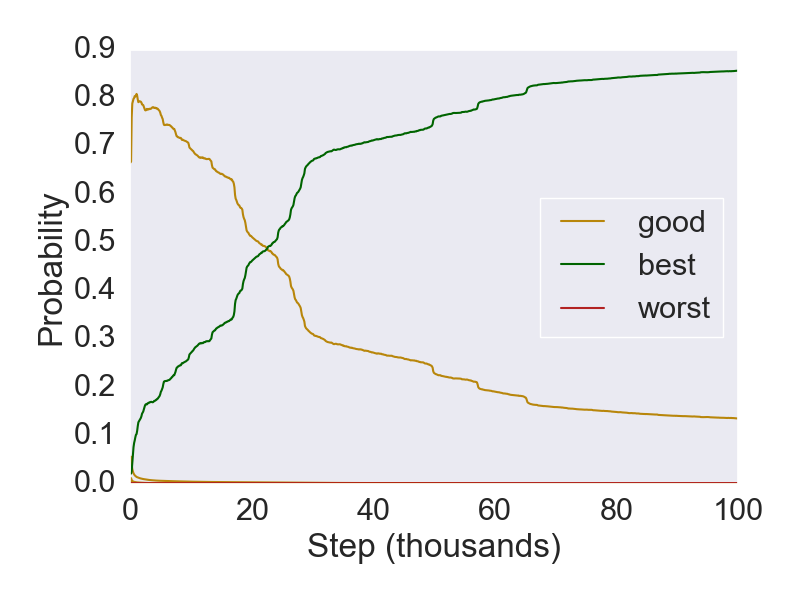}
			\caption{ \label{fig:reinforce-1_best2}}
		\end{subfigure}
		\begin{subfigure}[b]{0.33\textwidth} 
			\includegraphics[width=\columnwidth]{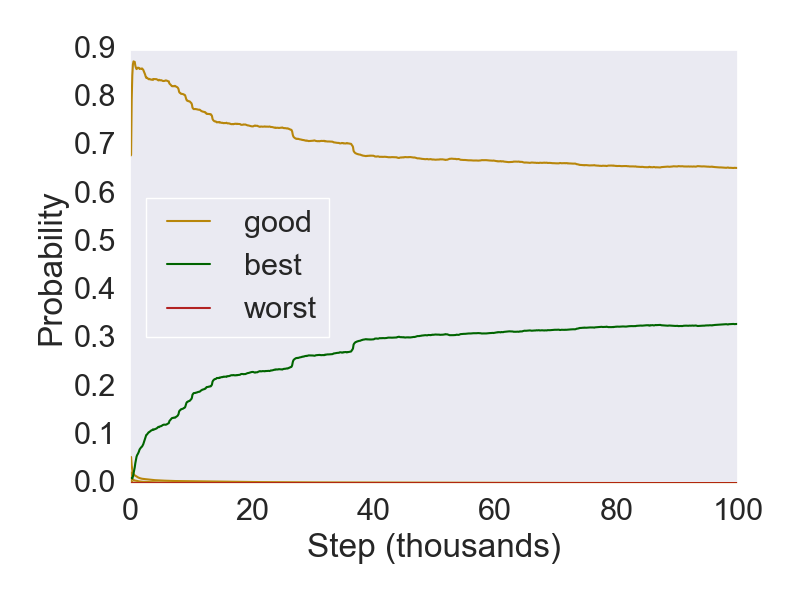}
			\caption{ \label{fig:reinforce-1_best3}}
		\end{subfigure}
		\caption{The probability of different tokens following {\sc Reinforce}, in the controlled simulations in the Constant Reward setting.
			The left/center/right figures correspond to simulations where the target token ($y_{best}$) was initially the second/third/fourth
			most probable token. The green line corresponds to the target token, yellow lines to medium-reward tokens and red lines to tokens with $r(y)=0$.  \label{fig:reinforce-1_3best}}
	\end{figure*}

	\section{NMT Implementation Details} \label{ap:technichalities}
	
	True casing and tokenization were used \cite{koehn2007koehn}, including escaping html symbols and "-" that represents a compound was changed into a separate token of =. Some preprocessing used before us converted the latter to \textit{\#\#AT\#\#-\#\#AT\#\#} but standard tokenizers in use process that into 11 different tokens, which over-represents the significance of that character when BLEU is calculated. BPE \cite{sennrich2016neural} extracted 30715 tokens.
	For the MT experiments we used 6 layers in the encoder and the decoder. The size of the embeddings was 512. Gradient clipping was used with size of 5 for pre-training (see Discussion on why not to use it in training). We did not use attention dropout, but 0.1 residual dropout rate was used. In pretraining and training sentences of more than 50 tokens were discarded. Pretraining and training were considered finished when BLEU did not increase in the development set for 10 consecutive evaluations, and evaluation was done every 1000 and 5000 for batches of size 100 and 256 for pretraining and training respectively. Learning rate used for rmsprop \cite{tieleman2012lecture} was 0.01 in pretraining and for adam \cite{kingma2015Adam} with decay was 0.005 for training.  4000 learning rate warm up steps were used. Pretraining took about 7 days with 4 GPUs, afterwards, training took roughly the same time. Monte Carlo used 20 sentence rolls per word.

	\section{Detailed Results for Constant Reward Setting}
\begin{figure}[h]
		\includegraphics[width=\columnwidth]{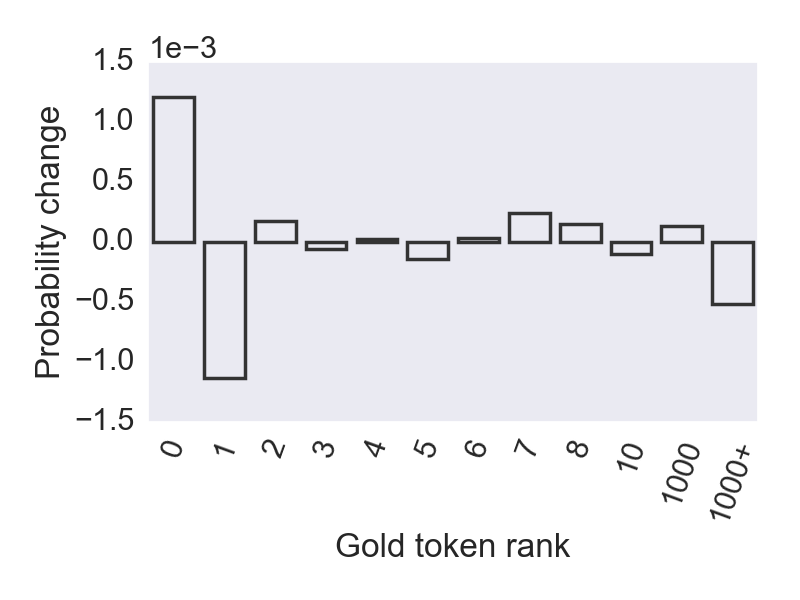}
		\caption{Difference between the ranks of $y_{best}$ in the reinforced with constant reward and the pretrained model.
			Each column $x$ corresponds to the difference in the probability that $y_{best}$ is ranked in rank $x$ in the reinforced 
			model, and the same probability in the pretrained model.\label{fig:constant_rank_diffs}}
	\end{figure}

	We present graphs for the constant reward setting in Figures \ref{fig:constant_rank_diffs} and \ref{fig:reinforce-1_3best}. Trends are similar to the ones obtained for the Simulated Reward setting.
	
\end{document}